**Title: Experimental Assessment of a Multi-Class AI/ML Architecture for Real-Time Characterization of Cyber Events in a Live Research Reactor**

**Authors**

Zachery Dahm, zdahm@purdue.edu, School of Nuclear Engineering, Purdue University, West Lafayette, IN 47907

Konstantinos Vasili, vasilik@purdue.edu, School of Nuclear Engineering, Purdue University, West Lafayette, IN 47907

Vasileios Theos, vtheos@purdue.edu, School of Nuclear Engineering, Purdue University, West Lafayette, IN 47907

Konstantinos Gkouliaras, kgkoulia@purdue.edu School of Nuclear Engineering, Purdue University, West Lafayette, IN 47907

William Richards, richa390@purdue.edu, School of Nuclear Engineering, Purdue University, West Lafayette, IN 47907

True Miller, mill1833@purdue.edu, School of Nuclear Engineering, Purdue University, West Lafayette, IN 47907

Brian Jowers, bjowers@purdue.edu, School of Nuclear Engineering, Purdue University, West Lafayette, IN 47907

Stylianos Chatzidakis, schatzid@purdue.edu, School of Nuclear Engineering, Purdue University, West Lafayette, IN 47907

**Title: Experimental Assessment of a Multi-Class AI/ML Architecture for Real-Time Characterization of Cyber Events in a Live Research Reactor**

**Authors**

Zachery Dahm[1], Konstantinos Vasili[1], Vasileios Theos[1], Konstantinos Gkouliaras[1], William Richards[1], True Miller[1], Brian Jowers[1], and Stylianos Chatzidakis[1,*]

[1]School of Nuclear Engineering, Purdue University, West Lafayette, IN 47907

*schatzid@purdue.edu

**Abstract**

There is increased interest in applying Artificial Intelligence and Machine Learning (AI/ML) within the nuclear industry and nuclear engineering community. Effective implementation of AI/ML could offer benefits to the nuclear domain, including enhanced identification of anomalies, anticipation of system failures, and operational schedule optimization. However, limited work has been done to investigate the feasibility and applicability of AI/ML tools in a functioning "live" nuclear reactor. Here, we go beyond the development of a single model and introduce a multi-layered AI/ML architecture that integrates both information technology and operational technology data streams to identify, characterize, and differentiate (i) among diverse cybersecurity events and (ii) between cyber events and other operational anomalies. Leveraging Purdue University's research reactor, PUR-1, we demonstrate this architecture through a representative use case that includes multiple concurrent false data injections and denial-of-service attacks of increasing complexity under realistic reactor conditions. The use case includes 14 system states (1 normal, 13 abnormal) and over 13.8 million multi-variate operational and information technology data points. The AI/ML architecture comprises three binary classifiers for layered characterization of system abnormalities. Multiple AI/ML algorithms were applied and systematically evaluated to assess their performance. Random Forest was identified as the optimal AI/ML model, achieving robust performance with an F1 score of 100% for most tests due to its explainability and resilience across various parameter settings. The study demonstrated the capability of AI/ML to distinguish between normal, abnormal, and cybersecurity-related events, even under challenging conditions such as denial-of-service attacks. Combining operational and information technology data improved classification accuracy but posed challenges related to synchronization and collection during certain cyber events. While results indicate significant promise for AI/ML in nuclear cybersecurity, the findings also highlight the need for further refinement in handling complex event differentiation and multi-class architectures.

## 1. Introduction

Recent advancements in computational capacity have enabled faster, more reliable real-time AI/ML monitoring systems, the evolution of intrusion detection systems is a recent example. However, developing an AI/ML architecture for cyber event characterization for a nuclear reactor system presents a far greater engineering challenge. Such a system must be guaranteed to not only detect but properly characterize, i.e., identify, localize, and distinguish, a large number of cyber scenarios without jeopardizing the reactor's safety. While widespread deployment of such AI/ML technologies remains years away, establishing the scientific foundation, ideally in a low-risk nuclear environment, such as PUR-1, represents a first step in assessing the reliability of AI/ML technologies for future nuclear plant applications.

Applications of AI/ML algorithms in the nuclear domain have primarily focused on the development of online monitoring tools (Coble et al., 2015; Ramuhalli et al., 2018), fault detection (Saeed et al., 2020), diagnostics (Hu et al., 2021; Yang and Kim, 2018), prognostics (Rivas et al., 2024), transient and parameter identification, and optimization tasks such as fuel loading patterns or fuel temperature predictions (Radaideh et al., 2021). The approaches employ computational and machine learning techniques including Bayesian statistical learning, Gaussian processes, artificial neural networks, and even deep learning methods such as convolutional neural networks (CNNs) and long short-term memory (LSTM) networks. Advanced AI/ML methods, such as autoencoders, have been applied to nuclear systems, for example, to detect elevated noise levels in a PWR (Mena et al., 2023) and to analyze data from physical control rod drives to detect multiple malfunction mechanisms (Xu et al., 2023). Most of these models are trained on synthetic data generated from system codes, e.g., MCNP and RELAP5, rather than experimental measurements. Limited work has been carried out on cyber event detection using experimental data (Zhang et al., 2019; Zhang, 2020a; Zhang et al., 2020b). (Zhang et al., 2019) performed a study using measurements from a cyber-physical two-loop forced flow system subjected to cyberattacks, including denial of service, data exfiltration, and false data injection. More recently, (Dahm et al., 2025) and (Vasili et al., 2025) analyzed experimental measurements from PUR-1 to demonstrate an explainable AI (XAI) framework combined with a data-driven general anomaly detection and reconstruction models to identify false data injections or replay attacks and isolate attack root causes. A similar approach is also used in (Park et al., 2022) but applied to a different set of anomalies. In parallel, non-AI/ML methods for false data detection have been proposed, such as watermarking (Mo and Sinopoli 2009) (Mo et al., 2015) or for active monitoring (Zhai & Vamvoudakis, 2021). While promising, non-AI methods rely on linearized mathematical methods or involve the modification of reactor process data. However, the analysis in all four of these works is restricted to operational technology (OT) data. Demonstrations to real-world nuclear environments with imperfect and noisy data have been limited (Sandhu et al., 2023) and significant challenges remain, including data acquisition, real-time processing, model accuracy, and the integration of multiple AI/ML models into a robust and reliable framework (Zhao et al. 2021; NUREG-2261; Hall and Agarwal, 2024).

In this work, we develop an automated multi-class cyber event characterization architecture using AI/ML and implement a use case in PUR-1 that integrates both OT and IT data with multiple system states. The use case examines events with system state transitions from normal to abnormal, including cyber events of increasing complexity. In total, eight events with physical manifestations were simulated, yielding 14 distinct system states captured in real-time through multivariate dynamic OT/IT signals. The dataset included (i) 13,400,000 OT and 156,750 IT normal datapoints and (ii) 418,080 OT and 19,800 IT abnormal datapoints. Using this dataset, multiple AI/ML algorithms were applied and systematically evaluated to assess their performance, identify limitations, and determine the key variables influencing characterization accuracy.

## 2. Background

Cyber-physical systems have several advantages, and their adoption could enhance the safe, secure, and economic operation of an industrial facility (Colombo et al., 2014), including current and future nuclear systems (Lin et al., 2021) (Dipty et al., 2021). Despite the obvious benefits, the possibility of introducing new unintended vulnerabilities, previously non-existent due to the analog-based and proprietary nature of operation, cannot be ruled out (IAEA, 2011a). This concern is reinforced by a steady rise in cyber-threats and cyber-attacks over recent years. Event information collected in several online databases, including in the Industrial Security Incident Database (ISID), indicates that the number of cyber incidents against cyber-physical systems worldwide has increased significantly and is on an upward trend since 2001 (Byres et al., 2007). The majority of these incidents are due to the use of legacy unpatched software, third-party networking, or unencrypted communications, and originate mostly from the Internet via malware (e.g., viruses, Trojan horses, and worms), even though there have been a few acts of insider sabotage (Byres et al., 2007).

Historically, industrial control systems have operated in isolated "air-gapped" environments and relied on proprietary software, hardware, and communications technology, rarely sharing information with systems outside the facility. Unauthorized acts to compromise these systems required specific knowledge of system architectures not publicly available and physical access to equipment. However, as systems now become more and more interconnected with the Internet, the risk of cyber-attacks has increased (Paganini, 2020). The nuclear industry has been a target of such cyber-attacks for the past 30 years, with many attacks aimed at gaining intelligence on control networks. Since the Stuxnet attack, many other incidents involved cyber-physical systems specifically designed to compromise these families of devices, including Duqu/Flame/Gauss (2011), Shamoon (2012), Havex (2013), Dragonfly (2014), Black Energy (2015), and Triton (2017). These attacks targeted nuclear plants, electric grids, dams, gas pipelines, water facilities, and industrial environments (Paganini, 2020). These events confirm that cyber-physical components are prime targets for both individual and nation-state actors. It is safe to say that cybersecurity and development of intrusion prevention and detection tools tailored to cyber-physical systems is expected to play an increasingly vital role in ensuring the prevention, detection and response to any unauthorized acts that could hurt the functions of a nuclear system.

A typical cyber-physical architecture is shown in Figure 1. This architecture is based on the Purdue model, a widely used reference model in industrial control systems (Li, 1994). The Purdue model came to define the standard for a cyber-physical architecture and shows how the typical elements of an architecture interconnect, dividing them into zones that contain information technology and operational technology systems (Zscaler, 2022). The architecture includes multiple zones (layers) with various equipment, components, controllers, and communication channels with various signal transmission requirements. Layers 0 to 3 represent the control network while Layer 4 is the outside business network. Layers 0 to 3 rely on limited computational resources. For example, an industrial grade proprietary PLC will have up to 512 MB SDRAM memory and a single CPU at 1 GHz (compared to enterprise computing resources that can be unlimited). Data sampling rates vary from 1 Hz to 1 kHz.

Components of a cyber-physical architecture include a central repository, supervisory control station, human-machine interface (HMI), remote terminal units, digital controllers, and field devices (ACM, 2015). To exchange information and configure commands, the HMI is connected to other application servers and engineering workstations via a communication network that includes control servers, long and short-range communication devices, remote terminal units (RTUs), field programmable gated arrays (FPGAs), and/or programmable logic controllers (PLCs). The collected data is viewed on computers at a central site (possibly in a remote location) by operators or automated supervisory tools that issue commands to field devices (IAEA, 2011b)). Potential risks associated with this architecture include: (i) greater complexity that can increase the attack surface and exposure to attackers, (ii) interconnected networks that can introduce common vulnerabilities and increased opportunities for DoS attacks, and (iii) two-way information flows that can increase the number of entry points and introduction of malicious code or compromised hardware.

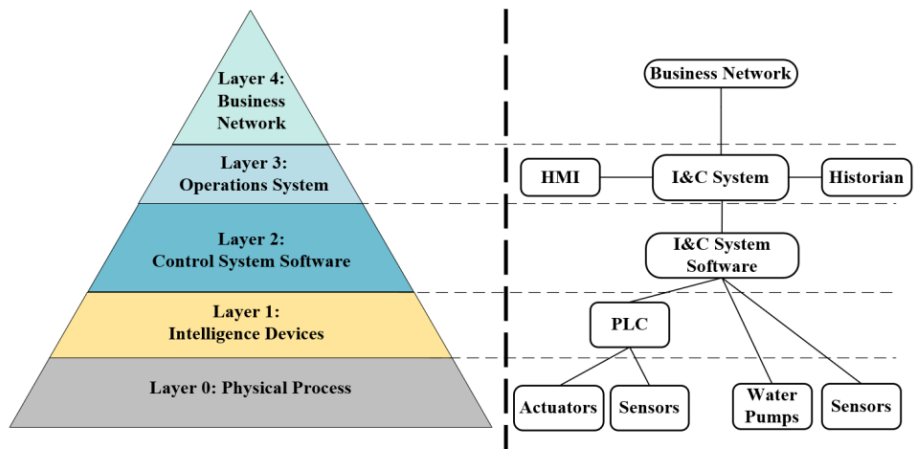

**Figure 1. Typical cyber-physical architecture (per Purdue model).**

To address cybersecurity concerns, cybersecurity guidance, regulations, and standards for protection of cyber-physical systems have been developed (and additional regulations, guidance, and standards are currently under development) and provided by several agencies (IAEA 2011a; IAEA 2011b; NIST 1998; NIST 2017; NIST 2018; NIST 2021). The U.S. NRC develops federal regulations and policies to ensure the safety and security of nuclear facilities. 10 CFR 73.54 (NRC, 2009) requires licensees to implement measures to ensure protection against cyber-attacks. Regulatory Guide 5.71 (NRC, 2010) provides guidance on how to implement and ensure compliance with 10 CFR 73.54. It recommends an architecture with different security levels, from high (level 4) to low (level 0) security requirements. Security levels define the protection degree required and the protection measures to be implemented. Levels 3 and 4 have the highest security requirements, aiming to ensure protection of safety systems from lower levels. Levels 3 and 4 allow only one-way outbound communications. Similar to U.S. NRC guidance, security levels are defined to group systems with similar security requirements. Security levels range from Level 1 (high security) to Level 5 (least protection), and no data are allowed to enter Level 1. IAEA also recommends a set of basic generic measures, e.g., intrusion prevention and detection tools, that apply to all levels, independent of security requirements (IAEA 2011a; IAEA 2011b).

A timeline of publicly disclosed cyber-incidents in nuclear and non-nuclear facilities and the current trend is shown in Figure 2. Notable cyber-incidents in nuclear and non-nuclear facilities and their causes and consequences are briefly summarized in Tables I and II. It is commonly acknowledged that the sophistication and frequency of cyber-incidents are increasing over time (Tounsi & Rais, 2018). A detailed examination of these events shows that certain types occur most frequently, including delay transmission, denial of service (DoS), eavesdropping, and false data transmission (FDI). Delay transmission events cause rapid increases in network traffic, which can slow device responses. DoS events consume all available device resources, potentially rendering them unresponsive to operational requests. Eavesdropping events compromise the confidentiality of communications, enabling unauthorized information gathering. FDI events involve the introduction of incorrect or modified data into devices or communication networks, which can disrupt normal system operation. These events can have different causes, e.g., inside or external actors seeking to compromise the system vs. equipment malfunction, but with symptoms that can be very similar. For example, the Davis-Besse and Browns Ferry incidents had different causes (malware vs. equipment malfunction), but both resulted in excessive network traffic and device unresponsiveness.

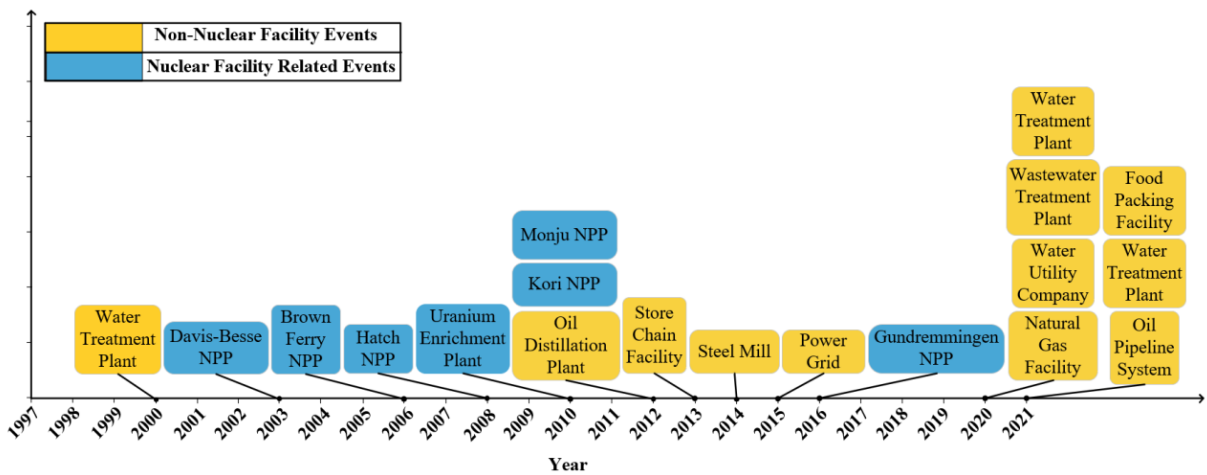

**Figure 2.** Timeline of cybersecurity incidents targeting critical infrastructure (Anton et al., 2017).

**Table I.** Publicly available cyber-incidents in nuclear facilities.

| Year | Location | Event Type | Cause | Consequences |
|------|----------|------------|-------|--------------|
| 2003 | Davis-Besse (Kim et al., 2020) | Delay transmission | Slammer Worm | Safety display inaccessible for 5h |
| 2006 | Browns Ferry (Poresky et al., 2017) | Denial of service | Equipment malfunction | Manual SCRAM by operator |
| 2008 | Hatch (Kim et al., 2020) | Denial of service | Maintenance update | Reactor SCRAM |
| 2010 | Enrichment Plant (Langner, 2011) | False data transmission | Malware | Equipment destroyed |
| 2014 | Monju (Poresky et al., 2017) | Eavesdropping | Malware | Data theft |
| 2014 | Kori NPP (Poresky et al., 2017) | Eavesdropping | Unknown | Data theft |
| 2016 | Gundremmingen NPP (Zhang, 2020) | Eavesdropping | Malware | Unknown |

**Table II.** Publicly available cyber-incidents in non-nuclear facilities.

| Year | Location | Event Type | Cause | Consequences |
|------|----------|------------|-------|--------------|
| 2000 | Water Treatment Plant (Hemsley & Fisher, 2018) | Sabotage | Insider | Release of untreated sewage |
| 2012 | Oil Distillation Plant (Bronk & Tikk-Ringas, 2013) | Denial of service | Malware | Data theft |
| 2013 | Store Chain Facility (Radichel & Northcutt, 2014) | Eavesdropping | Unknown | Data theft |
| 2014 | Steel Mill (Lee et al., 2014) | Unknown | Malware | Massive damage to the system |
| 2015 | Power Grid (Xiang et al., 2017) | Denial of service | Malware | Severe electric power outage |
| 2020 | Water Treatment Plant (Cimpanu, 2020) | Unknown | Unknown | Intrusion prevented |
| 2020 | Water Treatment Plant (Cimpanu, 2020) | Unknown | Unknown | Intrusion prevented |
| 2020 | Water Utility Company (Robles & Perlroth, 2021) | Sabotage | Unknown | Change of chemical level in the system |
| 2020 | Natural Gas Facility (US DOE, 2021) | Denial of Service | Malware | Facility shut down for two days |
| 2021 | Food Packing Facility (Batista et al., 2021) | Unknown | Malware | Operation system interruption |
| 2021 | Water Treatment Plant (CISA et al., 2021) | Denial of service | Malware | Manual operation by operators |
| 2021 | Oil Pipeline System (Hobbs, 2021) | Denial of service | Malware | Operation system interruption/data theft |

## 3. Use Case Overview

A use case encompassing a wide range of cyber events and OT/IT data types, capturing cyber-physical dynamics and events with physical process manifestations, was selected and implemented in PUR-1 to experimentally simulate adversary actions aimed at altering or manipulating physical processes. The initiating event is an unresponsive reactor trip, either as a result of a malfunction or as a result of a cyber event (i.e., unauthorized action of an adversary), implemented in combination with false data injections (i.e., modification of system objects) and/or DoS (i.e., modification of the relationship between system objects). The PUR-1 and remote OT/IT monitoring workstation which enables data collection in real time of more than 2000 parameters per sampling frequency are shown in Figure 3.

The use case is organized into five layers, each containing various objects and relationships between objects (NRC 2024, NRC 2024a-e). Objects (e.g., equipment, components, controllers) can be interconnected, and a change in one may affect another depending on the defined relationships (e.g., network connections, communication protocols, signal flow requirements), which can be either one- or two-way. For instance, a controller may use a two-way communication connection to exchange information with an actuator. Layers 0–3 are physically located at PUR-1, while Layer 4 represents the external network and is located in a separate room. Layers 3 and 4 communicate via Ethernet/TCP-IP, with IT security measures (e.g., firewalls, data diodes) separating them. Physical OT signals for training, validation, and testing of AI/ML algorithms are generated within Layers 0–3, while IT data originate from Layers 3 and 4. All data collection, computation, and analysis, including AI/ML training, validation, and testing, are performed at Layer 4, where an "attacker" PC is also connected via Ethernet to introduce cyber events (e.g., false data injection, denial of service). Within this framework, a mode describes the overall operational condition of the system (e.g., startup, power operation, standby, refueling, hot shutdown), while a state is a subset of a mode that reflects the system's specific condition. Any event affecting an object or its relationships results in a state change.

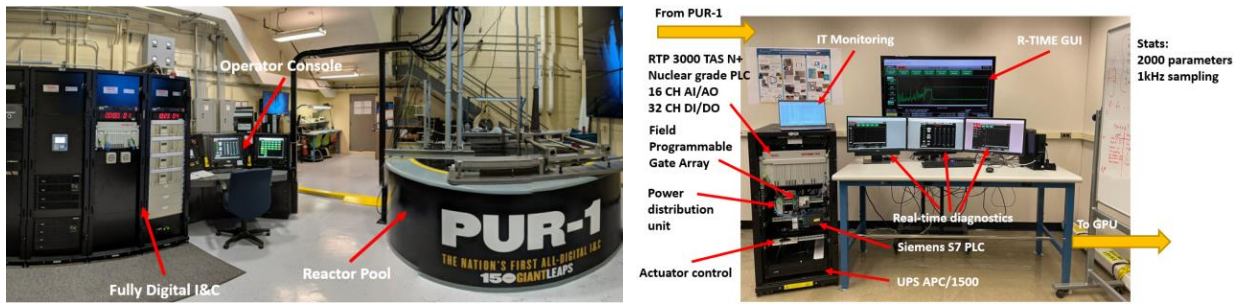

**Figure 3.** PUR-1 reactor with remote OT/IT monitoring workstation.

The implementation relies on several assumptions (NRC 2024, NRC 2024a-e). We assume adversary access to Layer 1 is possible but not Layer 0. This allows for events which may originate from any Layer except Layer 0 (see Figure 4 for Layer numbering and entry points). This constraint prevents modification or compromise of field sensors and aligns with the current limitations of the PUR-1 testbed, while still preserving system fidelity, since most Layer 0 devices in nuclear systems are analog-based; to our knowledge, no publicly available cyber event has originated from Layer 0. Second, to ensure adequate simplicity of the use case, we assume an adversary has no manipulation capability or access to the AI/ML architecture. This excludes scenarios in which the AI/ML architecture itself is compromised. Third, all AI/ML algorithms, training, and data processing are assumed to occur on a workstation located in Layer 4 without computing resource restrictions, reflecting the most probable location of such systems in practice. Finally, we assume that the AI/ML architecture receives identical information as the HMI, such that any false data injected at Layer 3 also propagates to Layer 4, with a one-way flow of data from Layer 3 to Layer 4 to maintain simplicity.

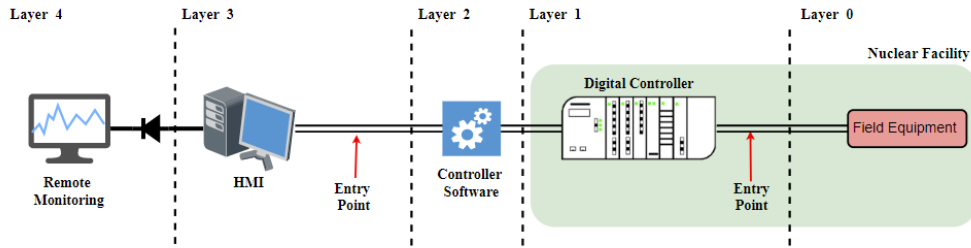

**Figure 4.** Example entry points from which cyber events may initiate.

When pressing the reactor trip button, during normal operation, magnet power to the control rods is cut instantly which results in full insertion in less than a second, reducing reactor power to residual decay heat. After a trip, sensors verify rod position to confirm system response. To evaluate this process, eight events were experimentally simulated: (1) manual trip with trip available, (2) manual trip with trip unavailable due to a cyber cause, (3) manual trip with trip unavailable due to a non-cyber malfunction, (4) FDI of one signal, (5) FDI of two signals, (6) FDI of three signals, (7) high-intensity DoS, and (8) low-intensity DoS. DoS was conducted using ethical penetration testing tools that generated continuous data packets, increasing latency and degrading or disabling system responsiveness, while FDI was implemented by altering operational data transmitted to a remote location. The use case includes all normal states, where the trip is available and reactor behavior remains acceptable. Abnormal states arise whenever the trip is unavailable—whether from cyber events, non-cyber malfunctions, or combined effects (e.g., DoS and FDI). FDI complexity was modeled incrementally: falsifying one signal (FDI #1), then two (FDI #2), and finally three (FDI #3), with each case a subset of the next. Simultaneous events are represented as overlapping states in Figure 5. Together, these events yielded 14 distinct system states: one normal and 13 abnormal (States 2–14). The event progression, state mapping, datasets, and affected objects are summarized in Figure 6. The number of potential states grows exponentially as event combinations increase.

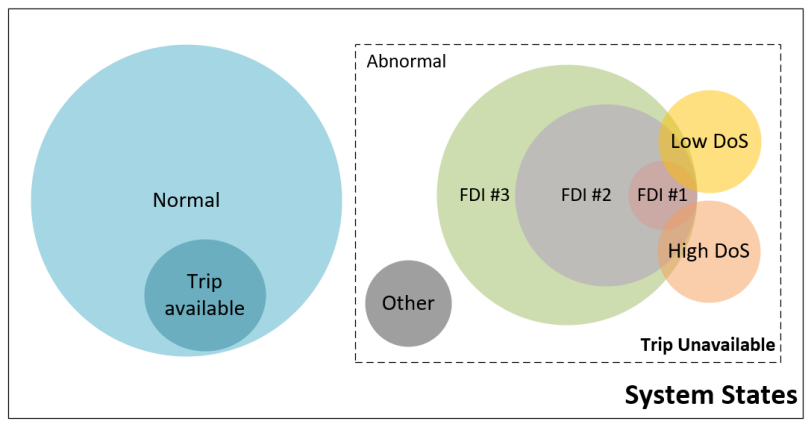

**Figure 5.** Implemented use case system states and events (adapted from NRC 2024).

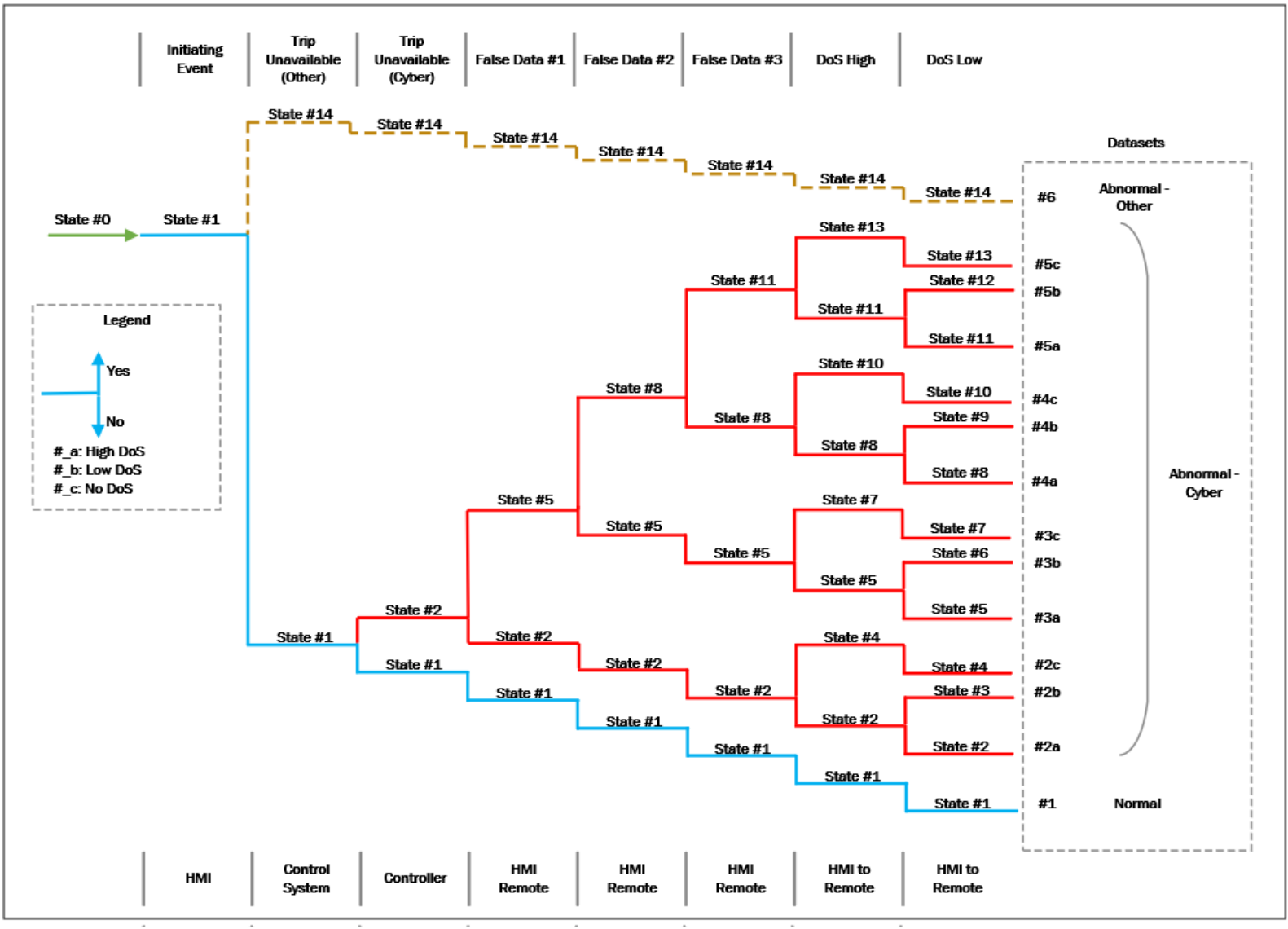

**Figure 6.** Use case event tree (adapted from NRC 2024).

**Data collection**

A total of 67 operational technology (OT) signals, selected from an available pool of approximately 2,000, and 11 information technology (IT) signals, selected from a pool of about 1,000, were collected based on their direct relevance to the use case. Signals that did not vary or were unrelated to the reactor trip process were excluded. The OT signals comprised physical measurements of process variables, alerts, alarms, commands, and status indications, while the IT signals represented network traffic, packet-level data, and host-related information such as resource utilization, services, and processes. All signals were recorded as dynamic numerical time series spanning multiple layers of the system. Other available data types, including metadata, configuration files, categorical variables, and text, were not included in this initial effort due to time limitations.

The collected data were organized into 14 datasets, one representing normal system operation and the remaining 13 corresponding to abnormal states. Dataset #1 described normal reactor behavior in the absence of cyber or physical disturbances. It contained 13.4 million datapoints from the 67 OT signals and 156,750 datapoints from the 11 IT signals, recorded between August 2022 and June 2023. This dataset captured a broad range of operational conditions, including reactor power levels spanning 0 to 100 percent and multiple reactor trips in cases where the trip function remained available. Such diversity was critical to representing the natural variability of the system. The corresponding IT data primarily captured network traffic between Layers 3 and 4, while variations in power, pool temperature, and operator-monitored signals during reactor trips are shown in Figure 7.

Abnormal datasets were generated to represent conditions in which the trip function was unavailable as a result of either cyber or non-cyber causes, as well as to simulate the effects FDI and DOS. Dataset #2a established a baseline abnormal condition where the trip function was unavailable due to a cyber event, in the absence of any additional disturbances. Since the specific attack mechanism was not analyzed, this condition was emulated by taking data from normal operation and manually altering the trip button indication from "0" to "1" every 20 seconds over a 1,560-second period, while leaving all other signals unmodified. The mismatch in trip button indication thus reflected the unavailability of the trip, as illustrated in Figure 7. Building on this baseline, additional abnormal datasets were produced by introducing DoS and FDI conditions. DoS scenarios were generated using Kali Linux penetration testing tools, which continuously injected packets between Layers 3 and 4 to increase latency and degrade system responsiveness. Under normal operation, network traffic averaged 30 packets per second; during low-intensity DoS this rate increased to approximately 870 packets per second, and during high-intensity DoS it reached 24,000 packets per second. While these conditions significantly altered IT data, they did not directly affect OT signals.

FDI scenarios were modeled by replacing true trip data with falsified data drawn from normal operation. For each case, reactor trip signals were modified by copying trip data from normal conditions into Dataset #2a, thereby obscuring the abnormal state. Each falsification spanned 120 seconds, covering 60 seconds before and after each trip indication, repeated across 13 reactor trips for a total falsification duration of 1,560 seconds per signal. Although the space of possible signal falsifications across 67 OT signals is vast, three signals displayed prominently on the operator's console were selected as representative adversary targets. FDI #1 falsified Channel 1 counts per second, FDI #2 falsified both Channel 1 counts per second and its change rate, and FDI #3 extended the falsification to Channel 2 counts per second. In this hierarchy, FDI #1 is a subset of FDI #2, and both are subsets of FDI #3. Each abnormal dataset contained 104,520 OT datapoints and 9,900 IT datapoints. Dataset #6 was defined as a relabeled version of Dataset #2a, serving to facilitate AI/ML training by explicitly associating signals with abnormal conditions. Figure 7 illustrates the differences between real and falsified signals across the FDI scenarios, while Figure 8 shows representative IT activity during low- and high-intensity DoS conditions.

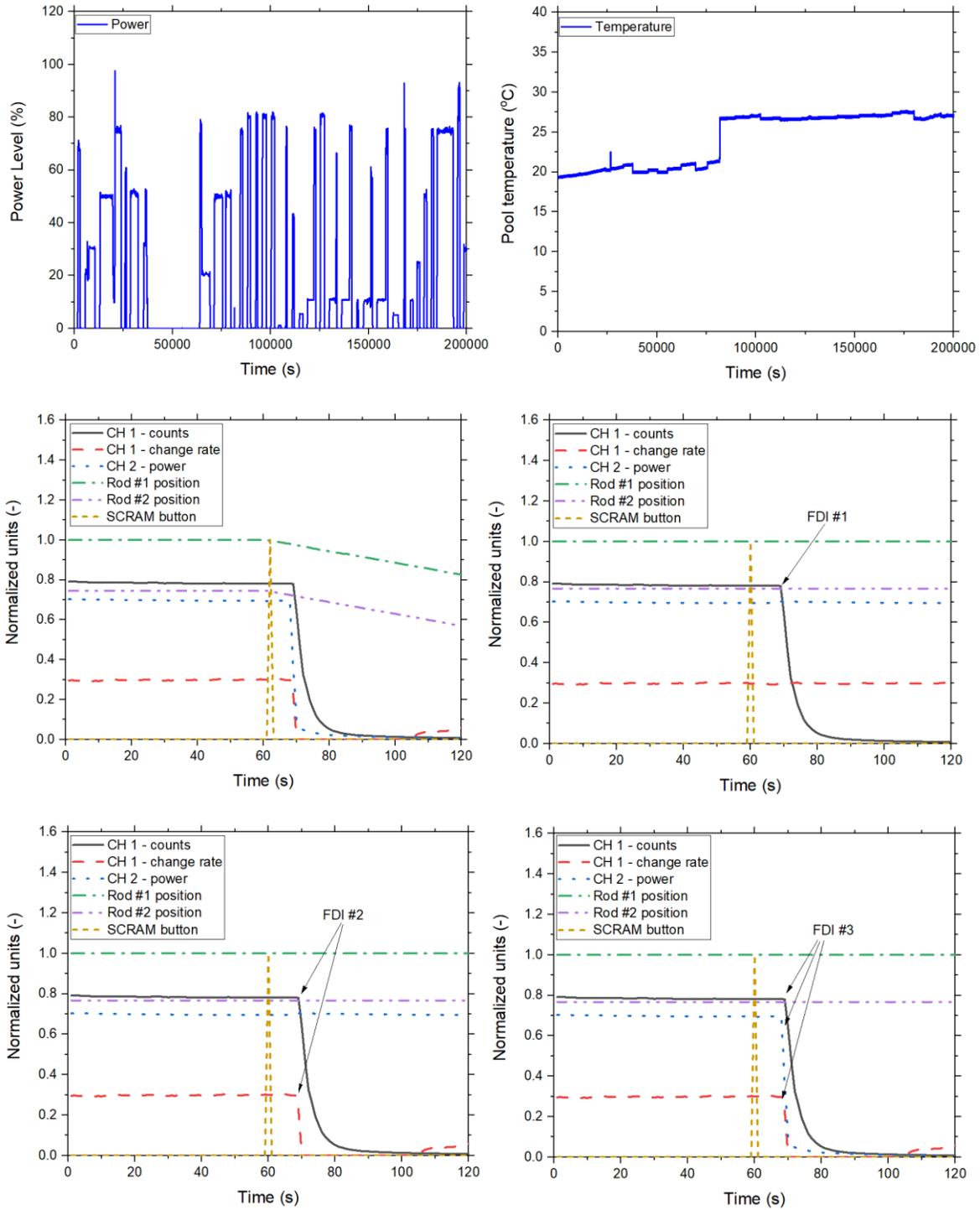

**Figure 7.** Normal power and temperature (top). Multiple signals during normal state (middle). Abnormal states 2 (middle left) and 5 (middle right). Abnormal states 8 (bottom left) and 11 (bottom right).

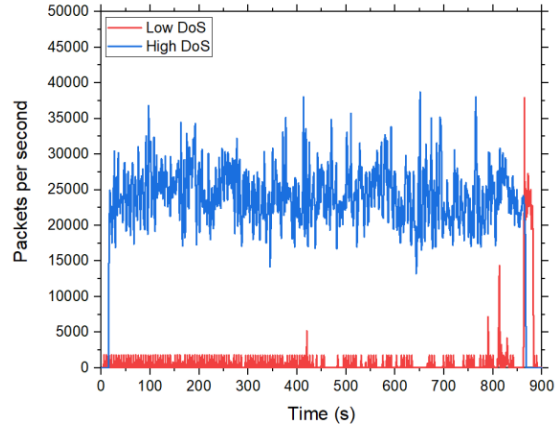

**Figure 8.** Network traffic for abnormal states 3 and 4.

**Data artifacts**

Artifacts were identified in the datasets from factors such as instrument response, connectivity delays, and operational variability. These artifacts were present across all datasets, though their frequency varied depending on the reactor state (e.g., shutdown versus operation). Such anomalies have the potential to impact the performance of AI/ML algorithms, and their presence underscores the need for data cleaning, preprocessing, and transformation prior to model development. Representative examples of signal fluctuations are shown in Figure 9. Three main categories of artifacts were observed. First, signal fluctuations attributable either to electronic noise or to natural variability in the underlying process (e.g., neutron flux). Second, outliers: a total of 109,660 outliers were identified in the OT data, representing 0.0482% of the 227 million datapoints collected prior to constructing the use case dataset. Of these, 98.02% (107,555 outliers) occurred during reactor operation, while 1.98% (2,173 outliers) occurred during shutdown. For context, operating data comprised approximately 13 million datapoints, or 5.6% of the total. Third, null values: 1,811,427 null entries were detected out of 227,637,525 total datapoints, corresponding to 0.78% of the collected data. Notably, all null values occurred during shutdown, with none observed during operation. No strong correlation was found between the occurrence of null values, outliers, or noise. Null values tended to appear in clusters, suggesting a possible link to network transmission protocols during data extraction rather than sensor malfunction. In addition to these systematic artifacts—fluctuations, outliers, and null values—sporadic effects linked to instrumentation and operator actions were also observed.

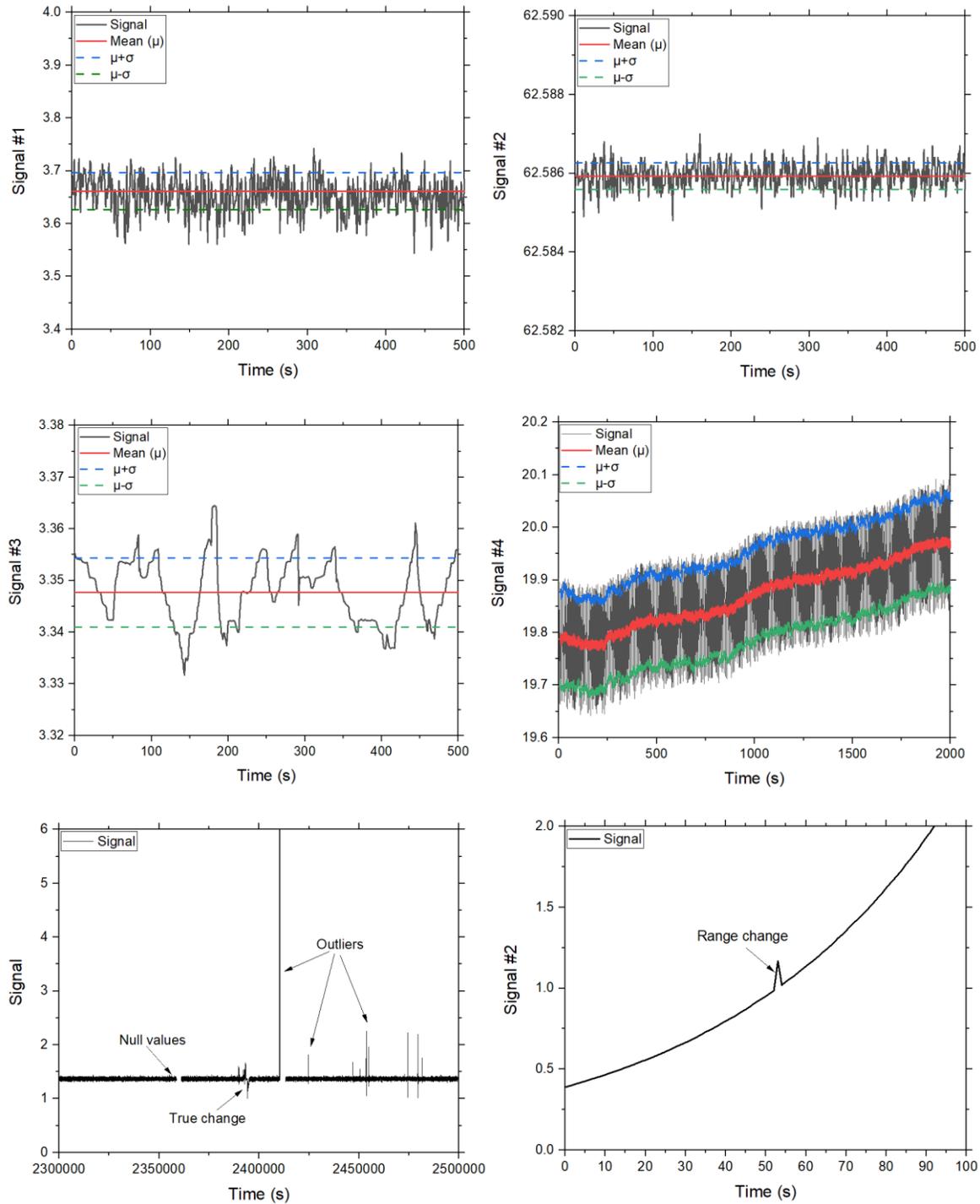

**Figure 9.** Signals with electronic noise (top). Signals with fluctuations due to natural variability in the data (middle left) or due to both electronic noise and fluctuations (middle left). Signals with null values and outliers (bottom left) and instrument-related artifact (bottom right).

**AI/ML Implementation**

A three-level classifier architecture was designed to demonstrate the potential of AI/ML for characterizing cyber events. Objectives include identifying the type of cyber event, distinguishing cyber events from other abnormal events, and further differentiating among specific cyber event types (Figure 10). Each level of the architecture performs an independent binary classification (0 = negative, 1 = positive), with its own AI/ML model trained, validated, and tested on a distinct dataset. By combining the outputs of all three levels, the classifier provides richer information than a single binary model, including the system state (normal or abnormal), whether the abnormality is cyber-related, and the specific event type (e.g., FDI, DoS, or a combination). The framework processes OT and IT data separately to improve information extraction efficiency, though tests with combined OT/IT data were also conducted for comparison.

At Level 1, OT data are analyzed to classify the system state as either normal (0) or abnormal (1). Level 2 then processes IT data to check for abnormal behavior consistent with a DoS attack. If both Level 1 and Level 2 outputs are normal, the system is classified as normal, and Level 3 is not activated. However, if Level 1 identifies an abnormality, Level 3 further examines the OT data to determine whether the anomaly is due to an FDI attack or another non-cyber-related event. A positive FDI classification at this stage confirms a cyber event. Conversely, an "Other" classification indicates that the anomaly was not cyber in origin. If Level 2 is simultaneously abnormal, the system is flagged as experiencing a DoS. A truth table summarizing all possible classifier outputs is provided in Table III. For example, an output of [1 0 1] corresponds to an abnormal state caused by a cyber event involving FDI. An output of [1 1 0] indicates an abnormal state resulting from a DoS. An output of [0 0 0] reflects normal operation.

Data normalization was applied to ensure all signals were on a comparable scale. Without scaling, high-magnitude signals, such as Channel 1 counts (measured in neutron counts per second, often in the millions or billions), could dominate smaller yet equally informative signals. Normalization was performed individually for each signal using both min-max and standard scaling methods; the impact on model performance was found to be negligible. Time-series data were then flattened into model inputs by defining a window length and including all data points within that window as a single input.

Datasets were split into training, validation, and testing subsets using three ratios: 50/30/20, 60/20/20, and 70/10/20, to assess the effect on performance. Various input parameters potentially influencing model performance, independent of hyperparameters, were considered. These included: window length (number of timesteps per input), window step (number of consecutive points between windows), balance ratio of normal to abnormal data, normalization method (min-max or standard), algorithm choice (Decision Tree, Random Forest, Logistic Regression, Linear SVM, Naïve Bayes), and data split ratio. Table IV summarizes these parameters and their values.

The best-performing model for each classifier level was selected and further optimized via hyperparameter tuning (e.g., learning rate, tree depth) using grid or random search. More than 5,000 parameter combinations were evaluated, revealing that algorithm type, balance ratio, and window length had the largest effect on performance, while window step and data split had minimal impact. Final model evaluation on the testing dataset was conducted using accuracy, F1 score, precision, recall, confusion matrices, and ROC curves. The overall methodology is illustrated in Figure 11.

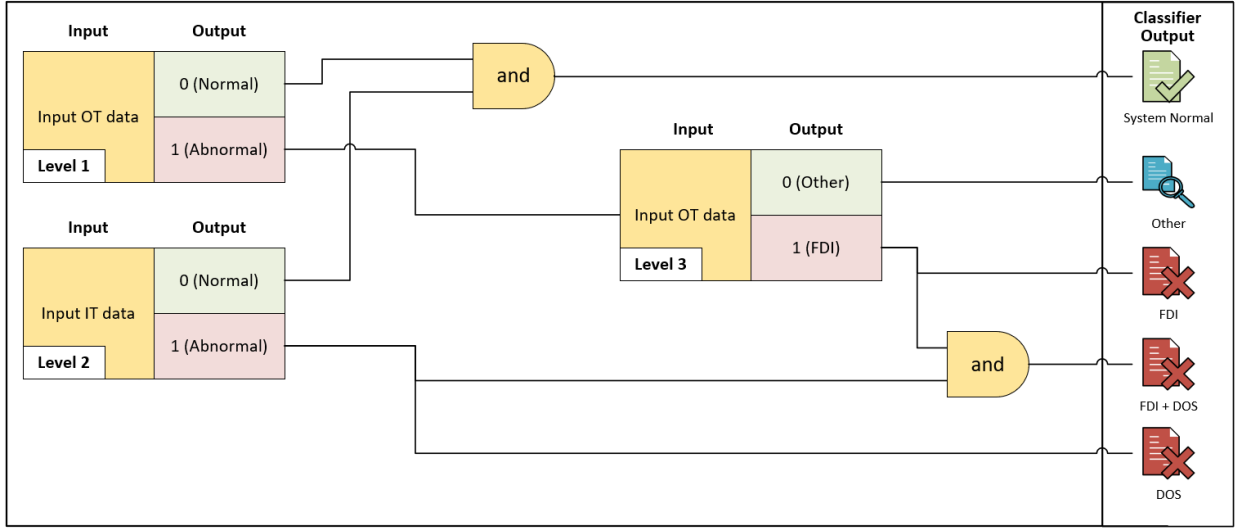

**Figure 10.** AI/ML architecture (adapted from NRC 2024).

**Table III.** Architecture truth table.

| Level 1 | Level 2 | Level 3 | Output | System Description | |
|---|---|---|---|---|---|
| | | | | State | Event Type |
| 0 | 0 | 0 | [0 0 0] | Normal state – no cyber events | |
| 1 | 0 | 0 | [1 0 0] | Abnormal | Other |
| 1 | 0 | 1 | [1 0 1] | Abnormal | Cyber - FDI |
| 0 | 1 | 0 | [0 1 0] | Abnormal | Cyber - DoS |
| 1 | 1 | 0 | [1 1 0] | Abnormal | Cyber - Other + DoS |
| 1 | 1 | 1 | [1 1 1] | Abnormal | Cyber - FDI + DoS |
| 0 | 0 | 1 | [0 0 1] | Not applicable | |
| 0 | 1 | 1 | [0 1 1] | Not applicable | |

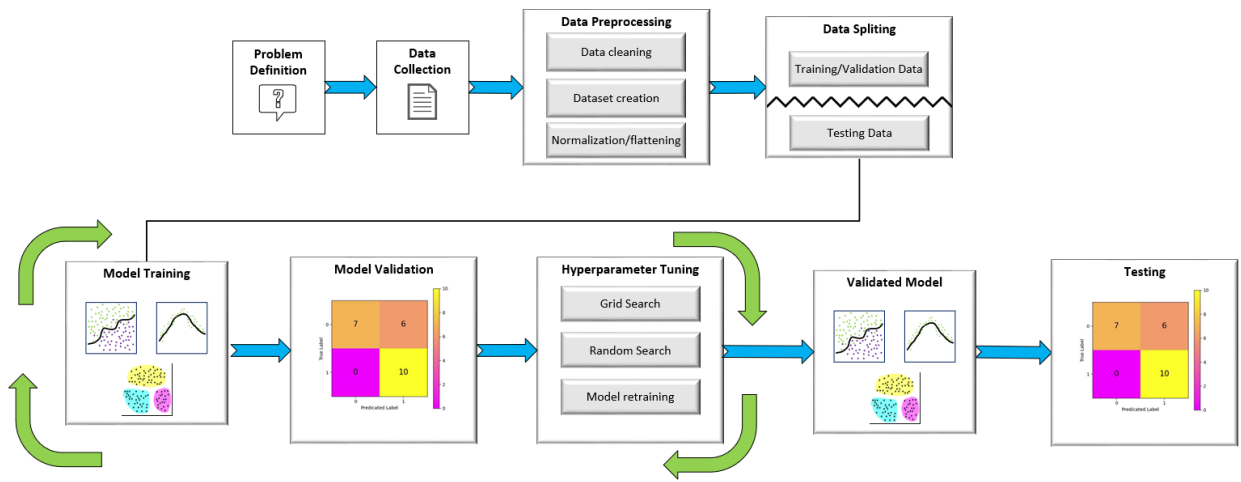

**Figure 11.** AI/ML implementation methodology (adapted from NRC 2024).

**Table IV.** Parameters used for identifying the best performing AI/ML model.

| Independent Variable | Description | Values |
|---|---|---|
| Window Length | Timesteps fed as input to the model | All levels: 1, 5, 10, 20, 30 |
| Window Step | Number of points between consecutive time windows | All levels: 1, 3, 5, 7, 10 |
| Training Balance Ratio | Normal to abnormal data ratio | Level 1: 1, 3, 5, 10, 20, 30<br>Level 2: 1, 3, 5, 10, 20, 30<br>Level 3: 0.33 |
| Validation Balance Ratio | Normal to abnormal data ratio | Level 1: 30<br>Level 2: 30<br>Level 3: 0.33 |
| Testing Balance Ratio | Normal to abnormal data ratio | Level 1: 30<br>Level 2: 30<br>Level 3: 0.33 |
| Scaling | Type of normalization applied to data | All levels: Min-max, standard |
| Algorithm | AI/ML algorithm used | All levels: Decision Tree, Random Forest, Logistic Regression, Linear SVM, Naïve Bayes |
| Train/Validation/Test Split | Dataset percent split | All levels: 50/30/20, 60/20/20, 70/10/20 |

Figure 12 illustrates the impact of different train/validation/test splits for two training balance ratios. Overall, the split had minimal effect on F1 scores for most models, except for SVMs at low training balance ratios, suggesting that the available training data were sufficient for effective learning. Higher training balance ratios, which provide more normal data, improved performance for SVMs and Logistic Regression, indicating these algorithms are more sensitive to dataset size. Across all scenarios, Random Forest and Decision Tree consistently outperformed SVM, Logistic Regression, and Naïve Bayes, independent of the training balance ratio.

Figure 13 further examines the effect of training balance ratio. For Logistic Regression, performance increased as the ratio approached 10, but beyond this point, the F1 score declined, with false negatives rising while false positives dropped to zero. Similar, though less pronounced, trends were observed in other models. This occurs because an excessively large normal dataset can bias the model toward minimizing false positives at the expense of false negatives, which poses a higher risk in nuclear systems.

Figure 13b shows the scenario where the training balance ratio is fixed at 30 while the absolute size of abnormal data is varied from 0 to 100% relative to Figure 13a. Performance improves with larger datasets, as both normal and abnormal data provide more information for training, highlighting that a minimum amount of data is necessary to achieve reliable performance. Insufficient data can lead to substantial increases in false positives and false negatives, even when overall accuracy and F1 scores appear high.

The influence of window length on model performance was examined for two training balance ratios (BR=1 and BR=30), as shown in Figure 14. Overall, window length had minimal impact on F1 scores for most models, except for SVMs. Random Forest and Decision Tree exhibited robustness across different balance ratios and window lengths, with only minor variations in performance. Consistent with earlier results, higher balance ratios generally improved F1 scores. Figure 15 shows the computational time for each model with a 20-second window length across different balance ratios. Random Forest required substantially more training time than Decision Tree, and its training time scaled approximately linearly with the size of the dataset, e.g., 30 times more data increased training time roughly 30-fold. While an 800-second training time was acceptable for this study, such scaling could pose challenges in large nuclear systems with far more signals and data. Decision Tree, with similar performance but much lower computational cost, offers a practical alternative.

In Level 1, Random Forest consistently outperformed other models across all variables and demonstrated robustness to varying balance ratios and window lengths. For Levels 2 and 3, all models performed similarly, likely due to the clearer separation between classes at these levels compared to Level 1. An important advantage of Random Forest and Decision Tree is explainability, allowing identification of erroneous model outputs and the specific signals responsible, which is not feasible with SVM, Logistic Regression, or Naïve Bayes. However, Random Forest's longer training time can become limiting in high-dimensional scenarios, making Decision Tree a viable alternative that maintains explainability with lower computational cost.

Considering these factors, Random Forest was chosen as the best model for all levels, both when processing OT and IT datasets separately (Levels 1–3) and when combining OT and IT in a modified Level 1. The final configuration of best-performing parameters included a window length of 20 seconds, a window step of 1 second, standard scaling, a 60/20/20 train/validation/test split, and a training balance ratio of 20.

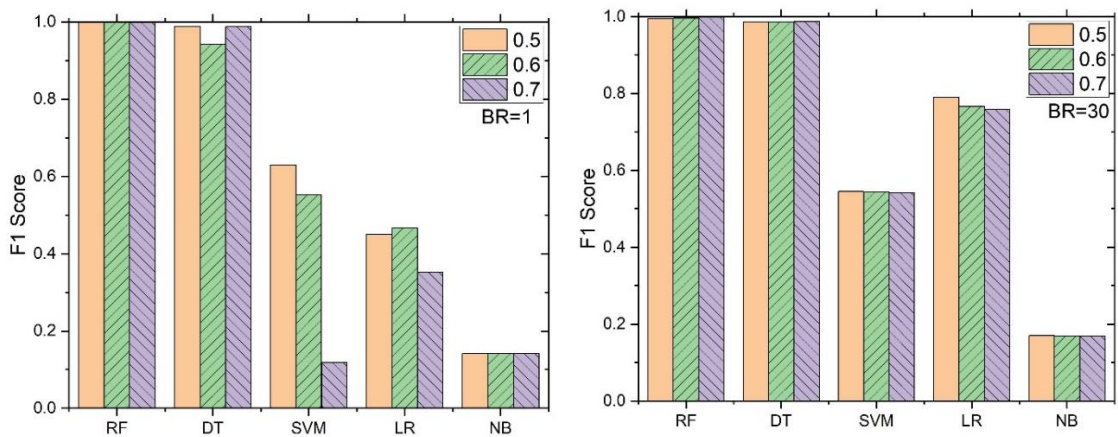

**Figure 12.** Effect of training data split (50%, 60%, and 70%) in AI/ML model performance for two training balance ratios (BR=1 and BR=30).

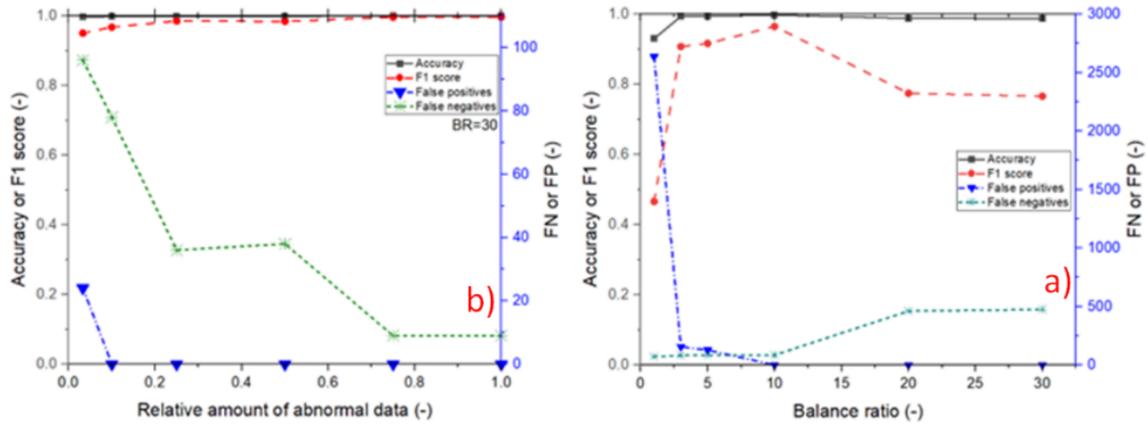

**Figure 13.** Effect of balance ratio in the Logistic regression model performance (left) and effect of changing the amount of abnormal data with constant balance ratio for Random Forest (right).

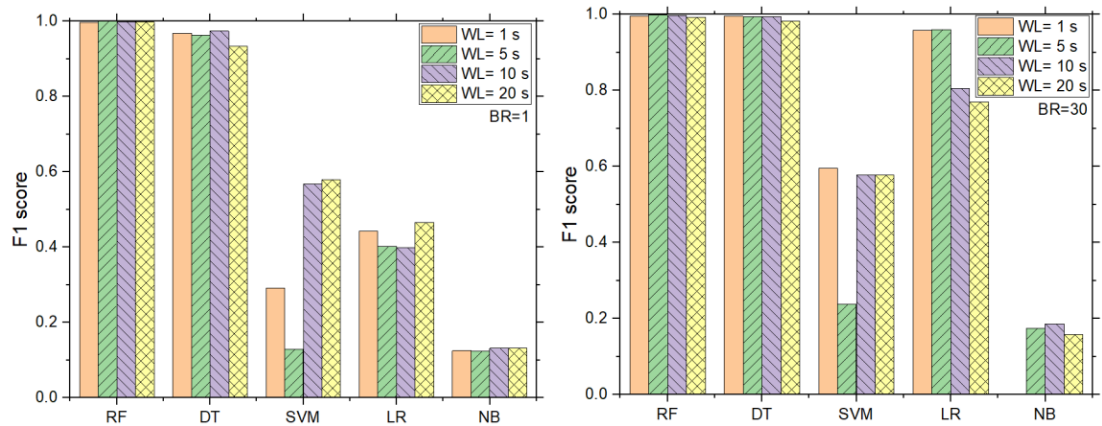

**Figure 14.** Effect of window length in AI/ML model performance for two different balance ratios.

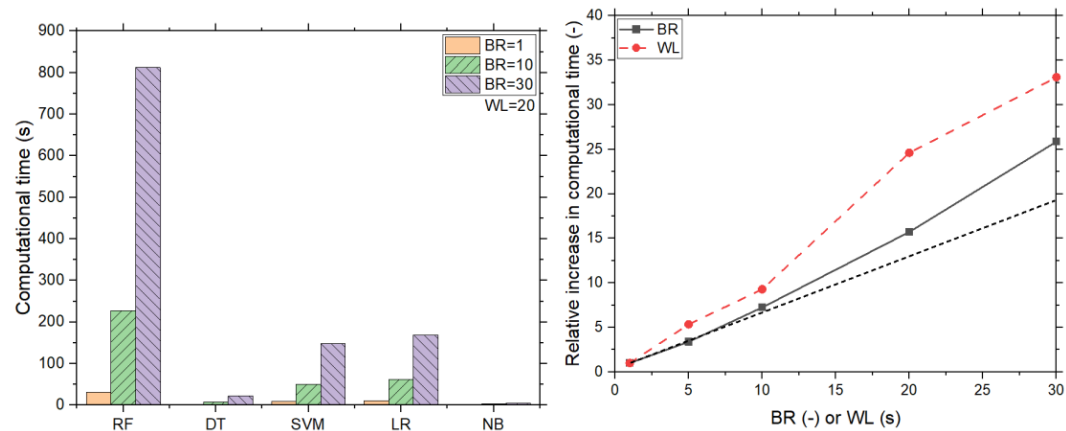

**Figure 15.** Computational time as a function of AI/ML model and training balance ratio (left). Relative increase in computational time for Random Forest as a function of training balance ratio and window length (right).

**Performance evaluation**

*Performance evaluation using separate OT and IT*

Using Random Forest models for all three classifier levels, each model was retrained with the optimized parameters and tested with a balance ratio of 30. As shown in Figure 16 and Table V, Levels 2 and 3 achieved perfect performance with an F1 score of 1.0, while Level 1 reached an F1 score of 99.7%, with six false negatives and no false positives. To further assess robustness, out-of-training datasets were created containing false data injections in Channels 3 and 4 combined with a DoS event, which were not present in the original training data. As shown in Figure 17 and Table VI, Level 1 identified abnormal events with an F1 score of 86% and 361 false negatives, Level 2 correctly identified DoS events with an F1 score of 100% and zero false positives or negatives, whereas Level 3 showed lower performance with an F1 score of 34% and 5,981 false positives.

*Performance evaluation using combined OT and IT*

OT and IT signals were combined to produce a dataset of 78 signals, which was used to train and test models on a binary classification task (normal vs. abnormal) using the optimal parameters. At a balance ratio of 1, Random Forest and Decision Tree achieved perfect F1 scores of 100% with no false positives or negatives, while SVM, Logistic Regression, and Naïve Bayes achieved F1 scores between 60% and 80% (Table VII, Figure 18). Although combining OT and IT signals allows classification of normal versus abnormal states, differentiating between multiple event types would require a multi-class architecture. Challenges in combining OT and IT data include differing data collection methods and sampling frequencies.

*Overall Classifier Performance*

The three-level classifier, combining Levels 1–3, functions as a multi-class classifier with six classes: Normal (0), Other/malfunction (1), FDI (2), DoS (3), Other + DoS (4), and FDI + DoS (5) (Table III). To evaluate performance, datasets representing each class were processed through the classifier, and the combined output was compared to the true class. Figure 19 shows the confusion matrix for all levels. The classifier correctly identified the majority of instances, with minimal misclassifications: Class 4 was misclassified once as Class 0, and Class 3 was misclassified once as Class 5. Overall, this demonstrates that combining multiple binary classifiers into a hierarchical architecture provides enhanced detection capabilities and improved performance relative to individual classifiers.

**Table V.** Performance metrics for separate OT and IT.

|         | **Accuracy** | **F1 score** | **Precision** | **Recall** |
|---------|--------------|--------------|---------------|------------|
| Level 1 | 0.999        | 0.997        | 1.0           | 0.995      |
| Level 2 | 1.0          | 1.0          | 1.0           | 1.0        |
| Level 3 | 1.0          | 1.0          | 1.0           | 1.0        |

|        | L1-Predicted | | L2-Predicted | | L3-Predicted | |
|--------|----------|----------|----------|----------|----------|----------|
|        | Positive | Negative | Positive | Negative | Positive | Negative |
| Actual Positive | 1246 | 6 | 355 | 0 | 930 | 0 |
| Actual Negative | 0 | 37560 | 0 | 10668 | 0 | 310 |

**Figure 16.** Confusion matrix for Level 1 (L1), Level 2 (L2) and Level 3 (L3) for separate OT and IT.

**Table VI.** Performance metrics for separate OT and IT and out-of-training data.

|         | Accuracy | F1 score | Precision | Recall |
|---------|----------|----------|-----------|--------|
| Level 1 | 0.988    | 0.867    | 1.0       | 0.765  |
| Level 2 | 1        | 1        | 1         | 1      |
| Level 3 | 0.2      | 0.34     | 0.2       | 1      |

|        | L1-Predicted | | L2-Predicted | | L3-Predicted | |
|--------|----------|----------|----------|----------|----------|----------|
|        | Positive | Negative | Positive | Negative | Positive | Negative |
| Actual Positive | 1179 | 361 | 557 | 0 | 1541 | 0 |
| Actual Negative | 0 | 29952 | 0 | 10668 | 5981 | 0 |

**Figure 17.** Confusion matrix for Level 1 (L1), Level 2 (L2), and Level 3 (L3) with out-of-training data.

**Table VII.** Performance metrics for combined OT and IT.

|     | Accuracy | F1 score | Precision | Recall |
|-----|----------|----------|-----------|--------|
| RF  | 1        | 1        | 1         | 1      |
| DT  | 1        | 1        | 1         | 1      |
| SVM | 0.589    | 0.693    | 0.553     | 0.930  |
| LR  | 0.828    | 0.807    | 0.920     | 0.718  |
| NB  | 0.718    | 0.608    | 1         | 0.4373 |

|        | Predicted | |
|--------|-----------|-----------|
|        | Positive  | Negative  |
| Actual / Positive | 1603 | 0 |
| Actual / Negative | 0 | 1603 |

**Figure 18.** Confusion matrix for combined OT and IT.

|  | Predicted | | | | | |
|--|---------|---------|---------|---------|---------|---------|
|  | Class 0 | Class 1 | Class 2 | Class 3 | Class 4 | Class 5 |
| Class 0 | 10324 | 0 | 0 | 0 | 0 | 0 |
| Class 1 | 0 | 87 | 0 | 0 | 0 | 0 |
| Class 2 | 0 | 0 | 3 | 0 | 0 | 0 |
| Class 3 | 0 | 0 | 0 | 5 | 0 | 1 |
| Class 4 | 1 | 0 | 0 | 0 | 256 | 0 |
| Class 5 | 0 | 0 | 0 | 0 | 0 | 346 |

**Figure 19.** Confusion matrix of overall classifier.

**Conclusions**

A comprehensive use case was implemented in PUR-1 to evaluate AI/ML capabilities for cyber event characterization in nuclear systems using both OT and IT data. The use case encompassed normal and abnormal system states, including events with physical manifestations and a combination of cyber events of increasing complexity. Data were collected from 67 OT and 11 IT signals across all system layers, generating 13.4 million OT and 156,750 IT datapoints for normal states, and 418,080 OT and 19,800 IT datapoints for abnormal states. Abnormal events included manual trip unavailability, false data injections of increasing complexity, and high- and low-intensity denial of service, resulting in 14 distinct system states.

Data preprocessing included normalization, flattening, and splitting into training, validation, and testing sets with multiple ratios evaluated. Independent input parameters such as window length, window step, and training balance ratio were systematically explored across five AI/ML algorithms: Random Forest, Decision Tree, Support Vector Machines (SVM), Logistic Regression, and Naïve Bayes. Extensive testing revealed that Random Forest consistently outperformed all other models, demonstrating robustness to variations in window length, balance ratio, and data split, while also providing explainability for identifying signals contributing to erroneous model behavior, a capability not achievable with other algorithms.

Separate OT and IT data were evaluated using a three-level classifier architecture. Level 1 classified system states as normal or abnormal using OT data, Level 2 identified abnormal IT events indicative of DoS, and Level 3 further characterized OT abnormalities as FDI or other events. With optimized parameters—window length 20 s, window step 1 s, standard scaling, 60/20/20 data split, and training balance ratio 20—Random Forest achieved near-perfect performance: Levels 2 and 3 reached an F1 score of 100%, while Level 1 achieved 99.7% with six false negatives and zero false positives. Out-of-training data tests including additional falsified signals and DoS events demonstrated robust performance in Levels 1 and 2, while Level 3 had reduced performance (F1 = 34%, 5,981 false positives).

When OT and IT signals were combined, Random Forest and Decision Tree achieved perfect classification (F1 = 100%, 0 false positives/negatives), whereas SVM, Logistic Regression, and Naïve Bayes ranged from 60–80% F1. While combined OT/IT data allow differentiation between normal and abnormal states, multi-class differentiation of specific event types remains more challenging due to differences in sampling rates, monitoring tools, and practical limitations during events like DoS, where OT data collection may be disrupted.

Overall, this study demonstrates that AI/ML methods, particularly Random Forest, can effectively characterize complex cyber-physical events in nuclear reactor systems. Random Forest's robustness, high accuracy, and explainability make it a promising candidate for future nuclear cybersecurity applications. Although challenges remain in integrating OT and IT data and differentiating event types under all conditions, these results provide a technical foundation for future research and potential deployment of AI/ML architectures in operational nuclear plants.

**Disclaimer**

This work is based in part on a technical report prepared for the U.S. Nuclear Regulatory Commission (ML24193A008). This work does not contain or imply legally binding requirements. Nor does this work establish or modify any regulatory guidance or positions of the U.S. Nuclear Regulatory Commission and is not binding on the Commission.

**Acknowledgements**

**References**

Anton, S. D., Fraunholz, D., Lipps, C., Pohl, F., Zimmermann, M., & Schotten, H. D. (2017). Two decades of SCADA exploitation: A brief history. In Proceedings of the 2017 IEEE Conference on Application, Information and Network Security (AINS) (pp. 98–104). IEEE. https://doi.org/10.1109/AINS.2017.8270432

Armando, C. W., Karnouskos. S., and Bangemann, T. (2014). "Towards the Next Generation of Industrial Cyber-Physical Systems." In *Industrial Cloud-Based Cyber-Physical Systems: The IMC-AESOP Approach*, edited by Armando W. Colombo, Thomas Bangemann, Stamatis Karnouskos, et al. Springer International Publishing. https://doi.org/10.1007/978-3-319-05624-1_1.

Batista, F., Hirtzer, M., & Dorning, M. (2021, June 1). All of JBS's U.S. Beef Plants Were Forced Shut by Cyberattack. *Bloomberg*. https://www.bloomberg.com/news/articles/2021-05-31/meat-is-latest-cyber-victim-as-hackers-hit-top-supplier-jbs?sref=CIpmV6x8&leadSource=uverify%20wall

Bronk, C., & Tikk-Ringas, E. (2013). Hack or Attack? Shamoon and the Evolution of Cyber Conflict. *SSRN Electronic Journal*. https://doi.org/10.2139/ssrn.2270860

Byres, E., Leversage, D., & Kube, N. (2007, May). Security Incidents and Trends in SCADA and Process Industries. *The Industrial Ethernet Book*.

Cimpanu, C. (2020, November 6). Israel government tells water treatment companies to change passwords. *ZDNet.* https://www.zdnet.com/article/israel-says-hackers-are-targeting-its-water-supply-and-treatment-utilities/

CISA, FBI, US EPA, & NSA. (2021). *Ongoing Cyber Threats to U.S. Water and Wastewater Systems*. www.fbi.gov/contact-us/field-offices

Coble, J., Ramuhalli, P., Bond, L., Hines, J.W., Upadhyaya, B. (2015). A Review of Prognostics and Health Management Applications in Nuclear Power Plants. Int J Progn Health Manag 16.

Dahm, Z., Theos, V., Vasili, K., Richards, W., Gkouliaras, K., & Chatzidakis, S. (2025). A one-class explainable AI framework for identification of non-stationary concurrent false data injections in nuclear reactor signals. *Nuclear Engineering and Design*, *444*, 114359. https://doi.org/10.1016/j.nucengdes.2025.114359

Dipty, Tripathi, Lalit Kumar, Anil Kumar Tripathi, and Amrita Chaturvedi. 2021. "Model Based Security Verification of Cyber-Physical System Based on Petrinet: A Case Study of Nuclear Power Plant." *Annals of Nuclear Energy,* 159, 108306. https://doi.org/10.1016/j.anucene.2021.108306.

Hall., A., Agarwal, V. (2024). "Barriers to adopting artificial intelligence and machine learning technologies in nuclear power." Progress in Nuclear Energy, Volume 175, 2024, 105295, ISSN 0149-1970. https://doi.org/10.1016/j.pnucene.2024.105295.

Hemsley, K. E., & Fisher, R. E. (2018). *History of Industrial Control System Cyber Incidents*. INL/CON-18-44411-Revision-2.

Hobbs, A. (2021). *The Colonial Pipeline Hack: Exposing Vulnerabilities in U.S. Cybersecurity*. SAGE Publications: SAGE Business Cases Originals. https://doi.org/10.4135/9781529789768

Hu, G., Zhou, T., & Liu, Q. (2021). Data-Driven Machine Learning for Fault Detection and Diagnosis in Nuclear Power Plants: A Review. *Frontiers in Energy Research*, *9*, 663296. https://doi.org/10.3389/fenrg.2021.663296

IAEA (2011a). *Computer Security at Nuclear Facilities*. International Atomic Energy Agency, IAEA Nuclear Security Series No. 17, Technical Guidance.

IAEA (2011b). *Core Knowledge on Instrumentation and Control Systems in Nuclear Power Plants*. IAEA Nuclear Energy Series No. NP-T-3.12.

Kim, S., Heo, G., Zio, E., Shin, J., & Song, J. gu. (2020). Cyber attack taxonomy for digital environment in nuclear power plants. *Nuclear Engineering and Technology*, *52*(5), 995–1001. https://doi.org/10.1016/j.net.2019.11.001

Langner, R. (2011). Stuxnet: Dissecting a cyberwarfare weapon. *IEEE Security and Privacy*, *9*(3), 49–51. https://doi.org/10.1109/MSP.2011.67

Lee, R., Assante, M. J., & Conway, T. (2014). German Steel Mill Cyber Attack. *ICS Defense Use Case*.

Li, H. (1994). *A Formalization and Extension of the Purdue Enterprise Reference Architecture and the Purdue Methodology*. Dissertation, Purdue University.

Linyu, L., Athe, P., Rouxelin, P. (2021). "Development and Assessment of a Nearly Autonomous Management and Control System for Advanced Reactors." *Annals of Nuclear Energy* 150 (January): 107861. https://doi.org/10.1016/j.anucene.2020.107861.

Mena, P., Borrelli, R. A., & Kerby, L. (2023). Detecting Anomalies in Simulated Nuclear Data Using Autoencoders. *Nuclear Technology*, *210*(1), 112–125. https://doi.org/10.1080/00295450.2023.2214257

Mo, Y., Weerakkody, S., & Sinopoli, B. (2015). Physical authentication of control systems: Designing watermarked control inputs to detect counterfeit sensor outputs. *IEEE Control Systems*, *35*(1), 93–109. https://doi.org/10.1109/MCS.2014.2364724

Mo, Y., and Sinopoli. B., (2009). "Secure Control against Replay Attacks." *2009 47th Annual Allerton Conference on Communication, Control, and Computing (Allerton)*, September, 911–18. https://doi.org/10.1109/ALLERTON.2009.5394956.

NRC (2009). Title 10, Code of Federal Regulations, Part 73.54: Protection of digital computer and communication systems and networks. U.S. Nuclear Regulatory Commission (NRC), U.S. Government Publishing Office. https://www.nrc.gov/reading-rm/doc-collections/cfr/part073/part073-0054.html

NRC (2010). Regulatory guide 5.71: Cyber security programs for nuclear facilities. U.S. Nuclear Regulatory Commission (NRC). https://www.nrc.gov/docs/ML0903/ML090340159.pdf

NRC (2024). *"Characterizing Nuclear Cybersecurity States Using Artificial Intelligence/Machine Learning." Technical Letter Report TLR-RES/DE-2024-03,* U.S. Nuclear Regulatory Commission (NRC), 2024. (ML23040A008).

NRC (2024a). *"Research Plan Development." Technical Letter Report TLR-RES/DE-2024-03a,* U.S. Nuclear Regulatory Commission (NRC), 2024. (ML23040A169).

NRC (2024b). *"Identification of a Representative Use Case." Technical Letter Report TLR-RES/DE-2024-03b,* U.S. Nuclear Regulatory Commission (NRC), 2024. (ML23062A349).

NRC (2024c). *"Identification of AI/ML Technologies." Technical Letter Report TLR-RES/DE-2024-03c,* U.S. Nuclear Regulatory Commission (NRC), 2024. (ML23102A182).

NRC (2024d). *"Use Case Implementation." Technical Letter Report TLR-RES/DE-2024-03d,* U.S. Nuclear Regulatory Commission (NRC), 2024. (ML24052A002).

NRC (2024e). *"Performance Evaluation and Gap Analysis." Technical Letter Report TLR-RES/DE-2024-03e,* U.S. Nuclear Regulatory Commission (NRC), 2024. (ML24193A007).

NIST (1998). *Information technology security training requirements: A role- and performance-based model* (NIST Special Publication 800-16). National Institute of Standards and Technology (NIST), U.S. Department of Commerce. https://doi.org/10.6028/NIST.SP.800-16

NIST (2017). *An introduction to information security* (NIST Special Publication 800-12, Rev. 1). National Institute of Standards and Technology (NIST), U.S. Department of Commerce. https://doi.org/10.6028/NIST.SP.800-12r1

NIST (2018). *Risk management framework for information systems and organizations: A system life cycle approach for security and privacy* (NIST Special Publication 800-37, Rev. 2). National Institute of Standards and Technology (NIST), U.S. Department of Commerce. https://doi.org/10.6028/NIST.SP.800-37r2

NIST (2021). *Enhanced security requirements for protecting controlled unclassified information: A supplement to NIST Special Publication 800-171* (NIST Special Publication 800-172). National Institute of Standards and Technology (NIST), U.S. Department of Commerce. https://doi.org/10.6028/NIST.SP.800-172

NIST (2024) "Glossary." National Institute of Standards and Technology (NIST), Computer Security Resource Center, https://csrc.nist.gov/glossary.

NUREG-2261 (2023). *Artificial intelligence strategic plan: Fiscal years 2023–2027* (NUREG-2261). U.S. Nuclear Regulatory Commission (NRC). https://www.nrc.gov/docs/ML2310/ML23105A011.pdf

Paganini, P. (2020, September 9). Basic Process Control Systems and Safety Instrumented System Devices. *INFOSEC*.

Park, J. H., Jo, H. S., Lee, S. H., Oh, S. W., & Na, M. G. (2022). A reliable intelligent diagnostic assistant for nuclear power plants using explainable artificial intelligence of GRU-AE, LightGBM and SHAP. *Nuclear Engineering and Technology*, *54*(4), 1271–1287. https://doi.org/10.1016/j.net.2021.10.024

Poresky, C., Andreades, C., Kendrick, J., & Peterson, P. (2017). *Cyber Security in Nuclear Power Plants Cyber Security in Nuclear Power Plants Cyber Security in Nuclear Power Plants: Insights for Advanced Nuclear Technologies*.

Radaideh, Majdi I., Isaac Wolverton, Joshua Joseph, et al. 2021. "Physics-Informed Reinforcement Learning Optimization of Nuclear Assembly Design." *Nuclear Engineering and Design* 372 (February): 110966. https://doi.org/10.1016/j.nucengdes.2020.110966.

Radichel, T., & Northcutt, S. (2014). *Case Study: Critical Controls that Could Have Prevented Target Breach*.

Ramuhalli, P., Coble, J., & Shumaker, B. (2018, June). Robust Online Monitoring Technologies for Nuclear Power Plant Sensors. *Transactions of the American Nuclear Society*.

Rivas, A., Kyriakos Delipei, G., Davis, I., Bhongale, S., & Hou, J. (2024). *A System Diagnostic and Prognostic Framework Based on Deep Learning for Advanced Reactors*. https://www.elsevier.com/open-access/userlicense/1.0/

Robles, F., & Perlroth, N. (2021, February 8). *https://www.nytimes.com/2021/02/08/us/oldsmar-florida-water-supply-hack.html*. The New York Times. https://www.nytimes.com/2021/02/08/us/oldsmar-florida-water-supply-hack.html

Saeed, Hanan A., Min-jun Peng, Hang Wang, and Bo-wen Zhang. 2020. "Novel Fault Diagnosis Scheme Utilizing Deep Learning Networks." *Progress in Nuclear Energy* 118: 103066.

Sandhu, Harleen Kaur, Saran Srikanth Bodda, and Abhinav Gupta. 2023. "A Future with Machine Learning: Review of Condition Assessment of Structures and Mechanical Systems in Nuclear Facilities." *Energies* 16 (6): 6. https://doi.org/10.3390/en16062628.

Tounsi, W., & Rais, H. (2018). A survey on technical threat intelligence in the age of sophisticated cyber attacks. In *Computers and Security* (Vol. 72, pp. 212–233). Elsevier Ltd. https://doi.org/10.1016/j.cose.2017.09.001

US DOE. (2021). Colonial Pipeline Cyber Incident. https://www.energy.gov/ceser/colonial-pipeline-cyber-incident

Vasili, K., Dahm, Z. T., Richards, W., & Chatzidakis, S. (2025). *An Unsupervised Deep XAI Framework for Localization of Concurrent Replay Attacks in Nuclear Reactor Signals*.

Xiang, Y., Wang, L., & Liu, N. (2017). Coordinated attacks on electric power systems in a cyber-physical environment. *Electric Power Systems Research*, *149*, 156–168. https://doi.org/10.1016/j.epsr.2017.04.023

Xu, Y., Cai, Y., & Song, L. (2023). Latent Fault Detection and Diagnosis for Control Rods Drive Mechanisms in Nuclear Power Reactor Based on GRU-AE. *IEEE Sensors Journal*, *23*(6), 6018–6026. https://doi.org/10.1109/JSEN.2023.3241381

Yang, Jaemin, and Jonghyun Kim. (2018). "An Accident Diagnosis Algorithm Using Long Short-Term Memory." *Nuclear Engineering and Technology*, International Symposium on Future I&C for Nuclear Power Plants (ISOFIC2017), vol. 50 (4): 4. https://doi.org/10.1016/j.net.2018.03.010.

Zhai, L., & Vamvoudakis, K. G. (2021). A data-based private learning framework for enhanced security against replay attacks in cyber-physical systems. *International Journal of Robust and Nonlinear Control*, *31*(6), 1817–1833. https://doi.org/10.1002/rnc.5040

Zhang, F., Kodituwakku, H.A.D.E., Hines, J.W., Coble, J. (2019). Multilayer Data-Driven CyberAttack Detection System for Industrial Control Systems Based on Network, System, and Process Data. IEEE Trans Industr Inform 15, 4362–4369. https://doi.org/10.1109/TII.2019.2891261

Zhang, F. (2020a). Nuclear power plant cybersecurity. In *Nuclear Power Plant Design and Analysis Codes: Development, Validation, and Application* (pp. 495–513). Elsevier. https://doi.org/10.1016/B978-0-12-818190-4.00021-8

Zhang, F., Hines, J.W., Coble, J.B. (2020b). A Robust Cybersecurity Solution Platform Architecture for Digital Instrumentation and Control Systems in Nuclear Power Facilities. Nucl Technol 206, 939–950. https://doi.org/10.1080/00295450.2019.1666599

Zhao, Xingang, Junyung Kim, Kyle Warns, et al. 2021. "Prognostics and Health Management in Nuclear Power Plants: An Updated Method-Centric Review with Special Focus on Data-Driven Methods." *Frontiers in Energy Research* 9: 696785.

Zscaler (2022). *What is the Purdue Model for ICS Security*? Zscaler. https://www.zscaler.com/resources/security-terms-glossary/what-is-purdue-model-ics-security